\documentclass[runningheads]{llncs}
\usepackage{latexsym}
\usepackage{graphicx}
\usepackage{multirow}
\usepackage{tabularx}
\usepackage{subcaption}
\usepackage{wrapfig}
\usepackage{svg}
\usepackage[T2A]{fontenc}
\usepackage[utf8]{inputenc}
\usepackage[russian,english]{babel}  
\usepackage[fixlanguage]{babelbib}
\usepackage{hyperref}

\begin{document}
\title{Automatic Aspect Extraction from Scientific Texts}
%
%
\author{Anna Marshalova\inst{1}\and
Elena Bruches\inst{1,2}
\and Tatiana Batura\inst{2}}
%
\authorrunning{A. Marshalova et al.}
\institute{Novosibirsk State University, Novosibirsk, Russia \and A.P. Ershov Institute of Informatics Systems, Novosibirsk, Russia}
%
%
\maketitle              
\begin{abstract}
Being able to extract from scientific papers their main points, key insights, and other important information, referred to here as aspects, might facilitate the process of conducting a scientific literature review. Therefore, the aim of our research is to create a tool for automatic aspect extraction from Russian-language scientific texts of any domain. In this paper, we present a cross-domain dataset of scientific texts in Russian, annotated with such aspects as Task, Contribution, Method, and Conclusion, as well as a baseline algorithm for aspect extraction, based on the multilingual BERT model fine-tuned on our data. We show that there are some differences in aspect representation in different domains, but even though our model was trained on a limited number of scientific domains, it is still able to generalize to new domains, as was proved by cross-domain experiments. The code and the dataset are available at \url{https://github.com/anna-marshalova/automatic-aspect-extraction-from-scientific-texts}.

\end{abstract}
\section{Introduction}

As the number of published research papers increases, it becomes more and more challenging to keep abreast of them. Therefore, there is a growing need for tools that can automatically extract relevant information from scientific texts.

Unfortunately, although Russian is among the languages most commonly used in science \cite{skvortsova2022russian}, there is only a sparse amount of tools for automatic information extraction from Russian scientific texts. To make matters worse, most of them focus on certain domains, e.g. medicine \cite{shelmanov2015information,loukachevitch2023nerel} or information technologies \cite{bakiyeva2020hybrid,bruches2020entity}.

To address this, our study aims to create a tool capable of extracting the main points of the research, which we refer to as aspects, from Russian scientific texts of any domain. It could be used to select, summarize, and systematize papers and is likely to make scientific literature reviewing more efficient.


The main contributions of this paper are summarized as follows:
\begin{enumerate}
    \item {we present a dataset containing abstracts for Russian-language scientific papers on 10 scientific domains annotated with 4 types of aspects: Task, Contribution, Method, and Conclusion;}
    \item {we provide an algorithm for automatic aspect extraction.}
\end{enumerate}

The dataset as well as the algorithm implementation are publicly available and may be of some use to other researchers.

The rest of the paper is organized as follows. Section \ref{sec:related work} gives some background information about the task of aspect extraction, points out the factors influencing the selection of aspects to identify, and reviews datasets and methods for both Russian and English that could be used for this task or similar tasks. Section \ref{sec:dataset} describes our dataset and explains why the above-mentioned 4 aspects were chosen in our work. Section \ref{sec:approach} outlines the details of our approach, based on using BERT, fine-tuned on our data. Section \ref{sec:experiments} describes our experiments and presents their results. Finally, section \ref{sec:discussion} discusses the principal findings and limitations of the study and suggests the broader impact of the work.

\section{Related work}
\label{sec:related work}

\subsection{Background}
\label{sec:related work background}

Pieces of information extracted from scientific papers do not have a conventional name. They are referred to as key-insights \cite{nasar2018information}, core scientific concepts, scientific artifacts \cite{hassanzadeh2014identifying}, or scientific discourse categories \cite{ronzano2015dr}. Like \cite{gupta-manning-2011-analyzing} and \cite{huang-etal-2020-coda} we will call them \textit{aspects} of a paper.

Moreover, there is currently no consensus on which information should be considered important and extracted from papers. We have discovered that the sets of aspects chosen by different researchers largely depend on the task and the data considered in their work.

Firstly, in most related works, aspect extraction is solved as a sentence classification or named-entity recognition (NER) task. This largely affects the aspects considered in the work. For example, studies devoted to NER tend to consider aspects expressed in short phrases, e.g. Method and Tool. In studies on sentence classification, higher-level aspects, such as Background, Goal and Conclusion, are usually considered. In order to take into account both types of aspects, we propose to identify as aspects phrases of any length within a sentence. A similar approach to text unit identification is used in the task of argumentative zoning \cite{teufel1999argumentative}, which, however, focuses on the argumentative structure of texts, which also influences the list of extracted types of information.

Secondly, the set of aspects depends on the size of the texts used. The selected texts can be both full papers and their abstracts. In our work, the latter are used, but even though abstracts reflect main aspects of papers, some information might still be missed, e.g. information about related work
 \cite{ronzano2015dr} and further work \cite{zhang2023automatic}. As a result, sets of aspects extracted from full paper texts usually include a larger number of aspects.

Finally, the aspects identified in papers depend on their scientific domain, as there are a number of domain-specific aspects. For example, papers on evidence-based medicine are traditionally structured according to the PICO criterion\footnote{ Population/Problem (P), Intervention (I), Comparison (C) and Outcome (O)}, which strongly influences the sentence categories identified in such texts \cite{boudin2010combining,kim2011automatic}. \cite{jain-etal-2020-scirex} studies papers on machine learning, so such entities as Dataset and Metric are proposed to be extracted. In this regard, the development of a set of aspects that can be used for different domains is rather challenging.

\subsection{Datasets}
\label{sec:related work datasets}
The dataset presented in this paper has no direct analogues, as currently there are no datasets designed specifically for aspect extraction from Russian scientific texts of different domains. Let us give a view of its closest analogues.

For Russian, there are datasets for NER and relation extraction from texts on various topics. These may include scientific texts, such as abstracts of papers \cite{bruches2020entity,loukachevitch2023nerel}, as well as texts of other genres, for example, medical records \cite{dudchenko2019extraction,gavrilov2020feature,kivotova2020extracting,nesterov-etal-2022-ruccon} or publications and forum entries from domain-specific social media \cite{blinov2022rumedbench,miftahutdinov2020biomedical,sirotina2019named}.

Datasets for information extraction from English-language scientific texts include the ones for sentence classification in medical abstracts \cite{boudin2010combining,dernoncourt-lee-2017-pubmed,huang-etal-2020-coda,kim2011automatic}, computer science abstracts \cite{gonccalves2020deep}, papers on computational linguistics \cite{gupta-manning-2011-analyzing,zhang2023automatic}, and computer graphics \cite{ronzano2015dr}. There are also datasets for NER and relation extraction from English scientific texts on computer science, materials science, physics \cite{augenstein-etal-2017-semeval}, and machine learning \cite{jain-etal-2020-scirex}. Finally, there are datasets for argumentative zoning, containing full papers \cite{teufel1999argumentative}.

\subsection{Methods}
\label{sec:related work methods}

Rule-based methods of information extraction from scientific texts usually involve extracting text segments containing certain keywords and n-grams or lexico-grammatical patterns \cite{gupta-manning-2011-analyzing,bakiyeva2020hybrid}. The main disadvantage of rule-based approaches is the effort required for the creation of rules and templates, which are also usually domain-specific. 

The most popular classical machine learning methods used for this task include Conditional Random Fields (CRF) \cite{kim2011automatic,hassanzadeh2014identifying,sirotina2019named} and Support Vector Machines (SVM) \cite{boudin2010combining,ronzano2015dr,shelmanov2015information}, trained on manually constructed features.

As for deep-learning methods, some researchers use convolutional neural networks and recurrent neural networks with pretrained static embeddings such as GloVe and word2vec \cite{dernoncourt-lee-2017-pubmed,dudchenko2019extraction,gonccalves2020deep,kivotova2020extracting}.

However, the state-of-the-art method for information extraction from scientific texts is based on pretrained masked-language models such as BERT \cite{yamada-etal-2020-sequential,shang-etal-2021-span} and SciBERT \cite{huang-etal-2020-coda,jain-etal-2020-scirex,zhang2023automatic}. Both BERT \cite{devlin-etal-2019-bert} and SciBERT \cite{beltagy-etal-2019-scibert} have Russian analogues: ruBERT \cite{kuratov2019adaptation} and ruSciBERT \cite{gerasimenko2022ruscibert}, the former being used by some researchers for NER in biomedical abstracts \cite{loukachevitch2023nerel} and texts on cybersecurity \cite{tikhomirov2020using}.

\section{Dataset creation}
\label{sec:dataset}

The created dataset contains 200 abstracts to Russian-language scientific papers of 10 domains: Medicine and Biology, History and Philology, Journalism, Law, Linguistics, Mathematics, Pedagogy, Physics, Psychology, and Computer Science. 

In these texts, we chose to identify 4 aspects: the task solved in the research, the authors' contribution, the employed methods, and the conclusion of the study. Examples\footnote{The texts are originally in Russian and were translated into English only to provide examples in the paper.} for each of the aspects are shown in Table \ref{tab:aspect list}.

\begin{table}
\caption{The list of aspects identified in our dataset with examples.}
\centering
\begin{tabular}{|l|p{0.8\textwidth}|}
\hline
\textbf{Aspect} &\textbf{Example}\\
\hline
\textbf{Task} & \textit{The paper is devoted to the task of <Task> improving the performance of small information systems </Task>.}\\
\hline
\textbf{Contribution} & \textit{<Contrib> ancient settlements, fortifications and protective walls of the Margaritovka basin have been studied </Contrib>.}\\
\hline
\textbf{Method} & \textit{Measurement of electrical and viscoelastic characteristics of erythrocytes was carried out  by the method of <Method> dielectrophoresis </Method>.}\\
\hline
\textbf{Conclusion} & \textit{The author comes to the conclusion that <Conc> bilateral unequal political alliances reflect local specifics </Conc>.}\\
\hline
\end{tabular}
\label{tab:aspect list}
\end{table}


As shown in the example of annotated text in Fig \ref{fig:manual_ru}, aspects can be nested, i.e. an aspect can be a part of another aspect.

The annotation was performed by two of the authors of this paper: a bachelor student of computational linguistics and a senior researcher, guided by the annotation instructions, which were compiled according to the semantic and syntactic features of the texts. All disagreements were discussed and resolved jointly. The inter-annotator agreement was evaluated with a strict F1 score \cite{hripcsak2005agreement}:
\begin{equation} \label{eq:f1-agreement}
F1 = \frac{2a}{2a+b+c}.
\end{equation}
In this equation, $a$ stands for aspects that were identified by both annotators, while $b$ and $c$ are aspects identified by only one of the annotators respectively.
Macro-averaged over all 4 aspects, the score measured 0.92.

\begin{figure}[h]
    \centering
    \includegraphics[width=0.8\textwidth]{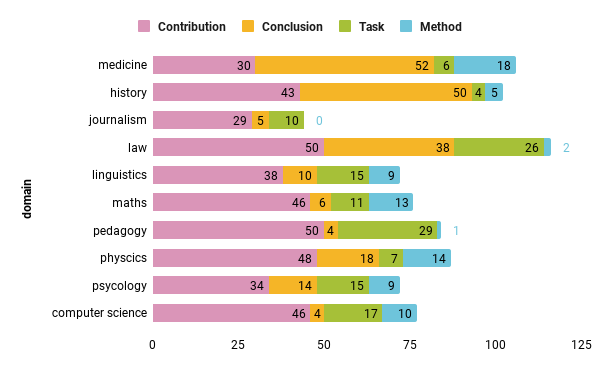}
    \caption{Distribution of aspect types over different domains in our dataset.}
    \label{fig:aspects_in_dataset}
\end{figure}

As a result, 836 aspects were identified, with half of them being the Contribution aspect (see Figure \ref{fig:aspects_in_dataset}). This is probably due to the fact that our dataset consists of abstracts, which are written to describe the authors' contribution. 

While length of one text is 115 tokens on average, it fluctuates across different domains from 50 to 177 tokens. Table \ref{tab:avg length vs aspects} shows that average text length positively correlates with the number of aspects identified in each domain.

\begin{table}
\caption{Average text length (tokens) in each domain compared to the number of aspects identified in these domains.}
\centering
\begin{tabular}{|l|r|r|}
\hline
\textbf{Domain}&\textbf{Average Text Length}&\textbf{Aspects Identified}\\
\hline
Journalism & 50 & 44 \\ 
\hline
Psychology & 86 & 72 \\ 
\hline
Linguistics & 90 & 72 \\ 
\hline
Pedagogy & 93 & 84 \\ 
\hline
Mathematics & 95 & 76 \\ 
\hline
Computer Science & 106 & 77 \\ 
\hline
Physics & 144 & 87 \\ 
\hline
Medicine & 146 & 106 \\ 
\hline
Law & 159 & 116 \\ 
\hline
History & 177 & 102 \\ 
\hline
\end{tabular}
    \label{tab:avg length vs aspects}
\end{table}

The number of aspects in one text varies from 0 to 13, and an average text contains 4 aspects. On average, one aspect consists of 12 tokens, but the length of an aspect largely depends on its type: Method and Task are similar to entities and are expressed in rather short phrases (3 -- 5 tokens), whereas Contribution and Conclusion are expressed in whole sentences or clauses (12 -- 16 tokens).

Contribution is sufficiently represented in all domains, while Conclusion prevails in medical texts since most of them describe the results of clinical studies. In addition, this aspect is often found in abstracts of papers on History, which often contain analysis of archaeological discoveries. Papers on Pedagogy and Mathematics are more practically oriented (they tend to propose new methods rather than describe experiments and observations); therefore, their abstracts contain fewer conclusions.

Methods are most often mentioned in texts on the natural and exact sciences: Medicine, Mathematics, Physics, and Computer science; these might also include Linguistics and Psychology.

As for the Task aspect, we discovered that in some domains, especially the humanities, we are not talking about tasks, but rather about problematic issues or objects of research (e.g. \textit{problems of dialogue}, \textit{self-identification issues} or even \textit{prepositions that function in journalistic discourse}). It was proposed to attribute them to the Task aspect as well.

\section{Automatic aspect extraction}
\label{sec:approach}
The pipeline of automatic aspect extraction performed by the presented tool is the following:
\begin{enumerate}
    \item \textbf{Text pre-processing.} At this step, we tokenize the texts using the Natural Language Toolkit (NLTK) library \cite{bird2009natural} (v3.6.2).
    \item \textbf{Aspect extraction processing.} At this step, we assign each token to corresponding aspects using the model described in section \ref{sec:approach model}.
    \item \textbf{Aspect post-processing.} At this step, we use some heuristics to improve aspect boundary detection. These involve, for example, removing one-word aspects and excluding punctuation and conjunctions from the end and the beginning of an aspect. After that, tokens assigned to one aspect mention are united into spans. Aspects expressed by nominal phrases are put in the nominative case using spaCy \cite{spacy2} (v3.5) and pymorphy2 \cite{korobov2015morphological} (v0.9.1). Finally, the spans are detokenized, and outputted in the format shown in Fig \ref{fig:extracted_ru}.
\end{enumerate}

\subsection{Model}
\label{sec:approach model}

We solve aspect extraction as a sequence labeling task. We do not employ BIO encoding, which is commonly used for sequence labelling, as in our annotation, aspects of the same type never follow each other in a row. They are separated by at least a comma or conjunction, so there is no need to distinguish between B-tags and I-tags, although we use O-tags for tokens that do not belong to any aspect. To take nested aspects into account, we use multilabel classification: a token can be labeled with one aspect or two aspects, if one of them is nested. Hence, the model is trained to label each token in a text with one or two of the four aspects (Task, Contribution, Method, Conclusion) or with the tag 'O'. 

\begin{figure}[h]
    \centering
    \includegraphics[width=0.8\textwidth]{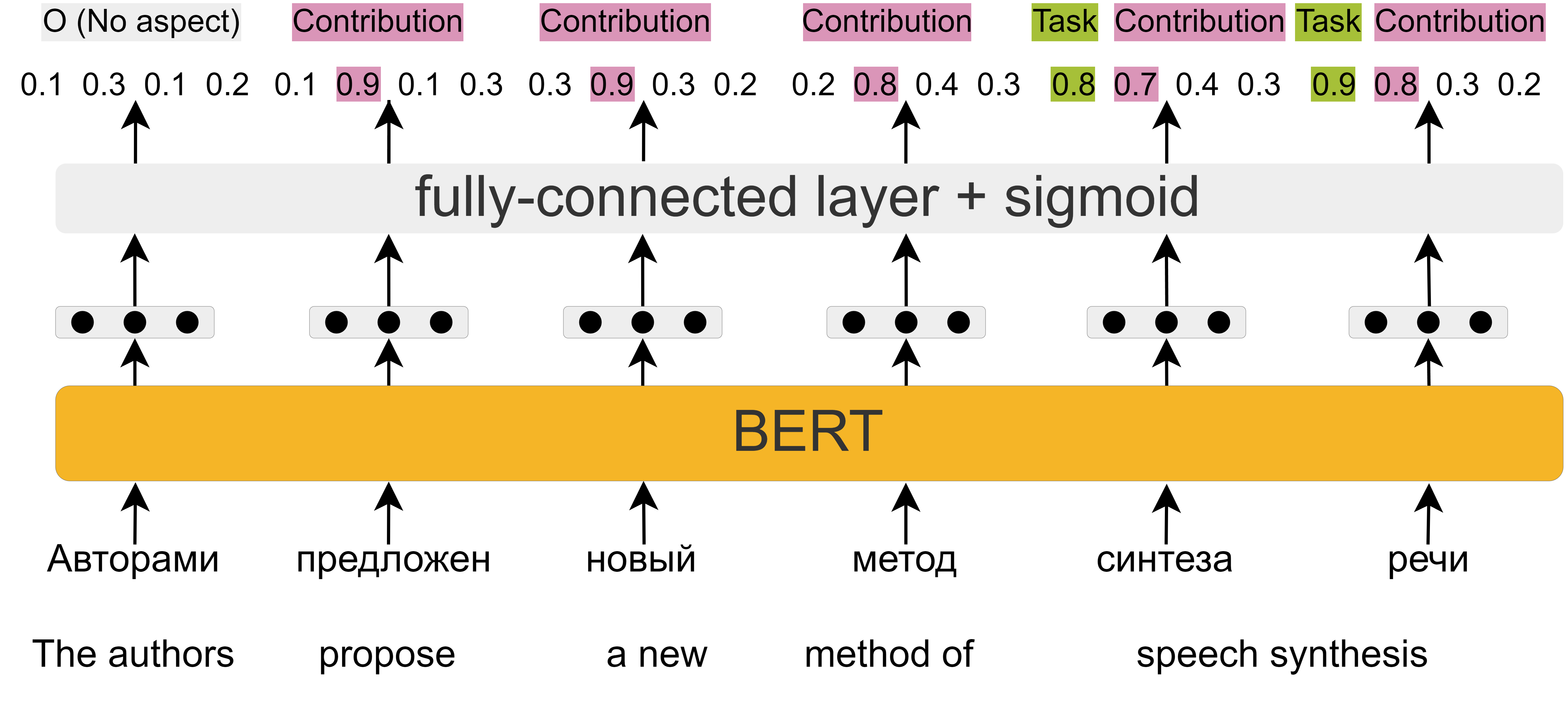}
    \caption{The scheme of the proposed model.}
    \label{fig:model_scheme}
\end{figure}

We use the model architecture shown in Fig. \ref{fig:model_scheme}. Namely, we use a pretrained BERT-like model fine-tuned on our dataset as an embedding layer and a fully-connected linear layer with a sigmoid activation function as a classifier. For each token, the model outputs an array of 4 values between 0 and 1, which can be interpreted as probabilities of assigning the token to each of the aspects. The token is labeled with up to two most probable aspects, whose probabilities are greater than the threshold of 0.5. If all of the values are less than the threshold, the token is labeled with the 'O' tag.

The model is trained with a batch size of 16, using binary cross-entropy as the loss function and Adam as the optimizer. The learning rate is set to $10^{-5}$.

We use multilingual BERT as the pretrained BERT-like model to fine-tune it on our data. This choice, as well as the choice of the model's architecture, is based on the results of our experiments, described in section \ref{sec:experiments choosing model}.

\section{Experiments and results}
\label{sec:experiments}

\subsection{Evaluation Method}
\label{sec:experiments evaluation}

For evaluation, we use precision, recall, and F1 for individual tokens, as well as Exact Match Ratio (EMR) of aspects --- the ratio of the number of correctly extracted aspects to the total number of aspects, which demonstrates the quality of aspect boundary detection. We calculate these metrics for each aspect and use macro-averaging to get the final scores.


To make evaluation more consistent, we employ 5-fold cross-validation. For each experiment, we provide mean and standard deviation (STD) of metrics across the folds.
Finally, to compare different models' performance, randomization tests \cite{noreen1989computer} are conducted.

\subsection{Experiments with models}
\label{sec:experiments choosing model}

To find which model would be the most suitable for our task, we conducted a number of experiments, using BERT-like models compatible with Russian: multilingual BERT (mBERT) \cite{devlin-etal-2019-bert}, ruBERT \cite{kuratov2019adaptation}, ruBERT-tiny2\footnote{\url{https://huggingface.co/cointegrated/rubert-tiny2}}, and ruSciBERT \cite{gerasimenko2022ruscibert}. The multilingual model performed best on our task (Table \ref{tab:pretrained model comparison}). However, the difference between the performance of mBERT and ruBERT turned out to be statistically insignificant (p = 0.3). Therefore, although we chose mBERT as the baseline model for further experiments, ruBERT could have also been chosen.

A series of experiments were also conducted with the multilingual BERT. These involve using CRF as a classifier, adding Bidirectional LSTM (BiLSTM) layer between the embedding and classification layers, and freezing the weights of the embedding layer to train only the classifier. However, these modifications did not outperform the baseline  model architecture with a fully-connected linear layer as a classifier (Table \ref{tab:model architecture comparison}). According to the randomization test, there is no statistically significant difference between the performance of the baseline model and the model with the BiLSTM layer (p = 0.28). However, training and inference are faster for the baseline model, so we consider it to be the best one.

\begin{table}[h]
\caption{Comparison of Macro F1 scores obtained with different models.}
\begin{subtable}[c]{0.5\textwidth}
\caption{Macro F1 scores obtained with different pretrained models.}
\centering
\begin{tabular}{|l|r|r|}
\hline
\multirow{2}{*}{\textbf{Model}} &  \multicolumn{2}{|c|}{\textbf{Macro F1}}\\
\cline{2-3}
&  \textbf{Mean} & \textbf{STD}\\
\hline
{mBERT} & \textbf{0.566} & 0.035\\
{ruBERT} & {0.546} & 0.071\\
rubert-tiny2 & {0.449} & 0.05\\
ruSciBERT & {0.287} & 0.066\\
\hline
\end{tabular}
\label{tab:pretrained model comparison}
\end{subtable}
\begin{subtable}[c]{0.5\textwidth}
\caption{Macro F1 scores obtained with different model architectures.}
\centering
\begin{tabular}{|l|r|r|}
\hline
\multirow{2}{*}{\textbf{Model}} &  \multicolumn{2}{|c|}{\textbf{Macro F1}}\\
\cline{2-3}
&  \textbf{Mean} & \textbf{STD}\\
\hline
{mBERT (baseline)} & \textbf{0.566} & 0.035\\
{mBERT + CRF} & 0.278 & 0.027\\
{mBERT + BiLSTM} & {0.54} & 0.023\\
{mBERT (frozen weights)} & 0.381 & 0.04\\
\hline
\end{tabular}
\label{tab:model architecture comparison}
\end{subtable}
\label{tab:model comparison}
\end{table}

Having compared the predictions of the models, we noticed that:
\begin{enumerate}
    \item RuSciBERT extracts spans of aspects that are too long, which leads to low precision.
    \item Most of the models extract nested aspects more often than needed and more often than the best model, which might mean that these models are less sure about their predictions, so they tend to assign tokens to two aspects. This also lowers the precision.
\end{enumerate}
These observations partially explain the obtained results. However, for a more detailed analysis of how the models work and to find out the reasons why mBERT outperforms specialised monolingual models such as ruSciBERT, additional experiments are needed, which we plan to perform in the future.

Table \ref{tab:best-metrcis} shows the metrics for the best model, which has the architecture shown in Fig \ref{fig:model_scheme} and uses multilingual BERT as an embedding layer. 

\begin{table}
\caption{Metrics for the best model after post-processing with heuristics.}
\centering
\begin{tabular}{|l|r|r|r|r|r|r|r|r|}
\hline
\multirow{2}{*}{\textbf{Tag}} &  \multicolumn{2}{c|}{\textbf{Precision}} &  \multicolumn{2}{c|}{\textbf{Recall}} &  \multicolumn{2}{c|}{\textbf{F1}} &  \multicolumn{2}{c|}{\textbf{EMR}}\\
\cline{2-9}
& \textbf{Mean} & \textbf{STD} &  \textbf{Mean} & \textbf{STD} &  \textbf{Mean} & \textbf{STD} &  \textbf{Mean} & \textbf{STD}\\
\hline
O (no aspect) & \textbf{0.735} & 0.019 & 0.602 & 0.101 & 0.658 & 0.055 & - & -\\
Task & 0.395 & 0.15 & 0.439 & 0.172 & 0.407 & 0.143 & 0.222 & 0.125\\
Contrib & 0.683 & 0.104 & \textbf{0.827} & 0.11 & \textbf{0.737} & 0.047 & \textbf{0.53} & 0.096\\
Method & 0.427 & 0.105 & 0.464 & 0.188 & 0.436 & 0.127 & 0.225 & 0.155\\
Conc & 0.58 & 0.132 & 0.617 & 0.066 & 0.593 & 0.094 & 0.225 & 0.09\\
\hline
Macro & 0.564 & 0.043 & 0.59 & 0.035 & 0.566 & 0.035 & 0.317 & 0.049\\
\hline
\end{tabular}
\label{tab:best-metrcis}
\end{table}

The best extracted aspect is Contribution, which can be explained by its frequency in the dataset. 
Exact match ratio measures lower than other metrics, indicating problems with aspect boundary detection.

\subsection{Cross domain experiments}
\label{sec:experiments cross-domain}
Our corpus contains texts on only a limited set of topics, while compiling a corpus that would cover all domains is a rather complicated task. We conducted 10 experiments in order to find out how our model performs on unseen domains. Within each of them, the texts of one of the domains were used to test a model trained on the texts of the rest 9 domains. For these experiments, 5-fold cross-validation was used to split data into train and validation.

The results of the experiments, shown in Table \ref{tab:cross-domain experiments}, are comparable with the best model's performance and confirm the ability of our model to generalize to various subject areas. 

\begin{table}
\caption{F1 scores for each of the cross-domain experiments. For each experiment, we used texts from one of the domains as a test set for a model trained on 9 other domains.}
\centering
\begin{tabular}{|l|r|r|}
\hline
\multirow{2}{*}{\textbf{Domain}} &  \multicolumn{2}{c|}{\textbf{Macro F1}}\\
\cline{2-3}
&  \textbf{Mean} & \textbf{STD}\\
\hline
Biology and Medicine & 0.471 & 0.038 \\
Computer Science & 0.543 & 0.053 \\
History and Philology & 0.45 & 0.015 \\
Journalism & 0.526 & 0.042 \\
Law & 0.473 & 0.02 \\
Linguistics & 0.472 & 0.013 \\
Math & 0.513 & 0.028 \\
Pedagogy & 0.529 & 0.066 \\
Physics & 0.536 & 0.038 \\
Psychology & 0.51 & 0.068 \\
\hline
\end{tabular}
\label{tab:cross-domain experiments}
\end{table}

\subsection{Analysis of automatic annotation}
\label{sec:experiments analysis}

\begin{figure*}[h]
    \begin{subfigure}[t]{0.5\textwidth}
    \centering
    \includegraphics[width=\textwidth]{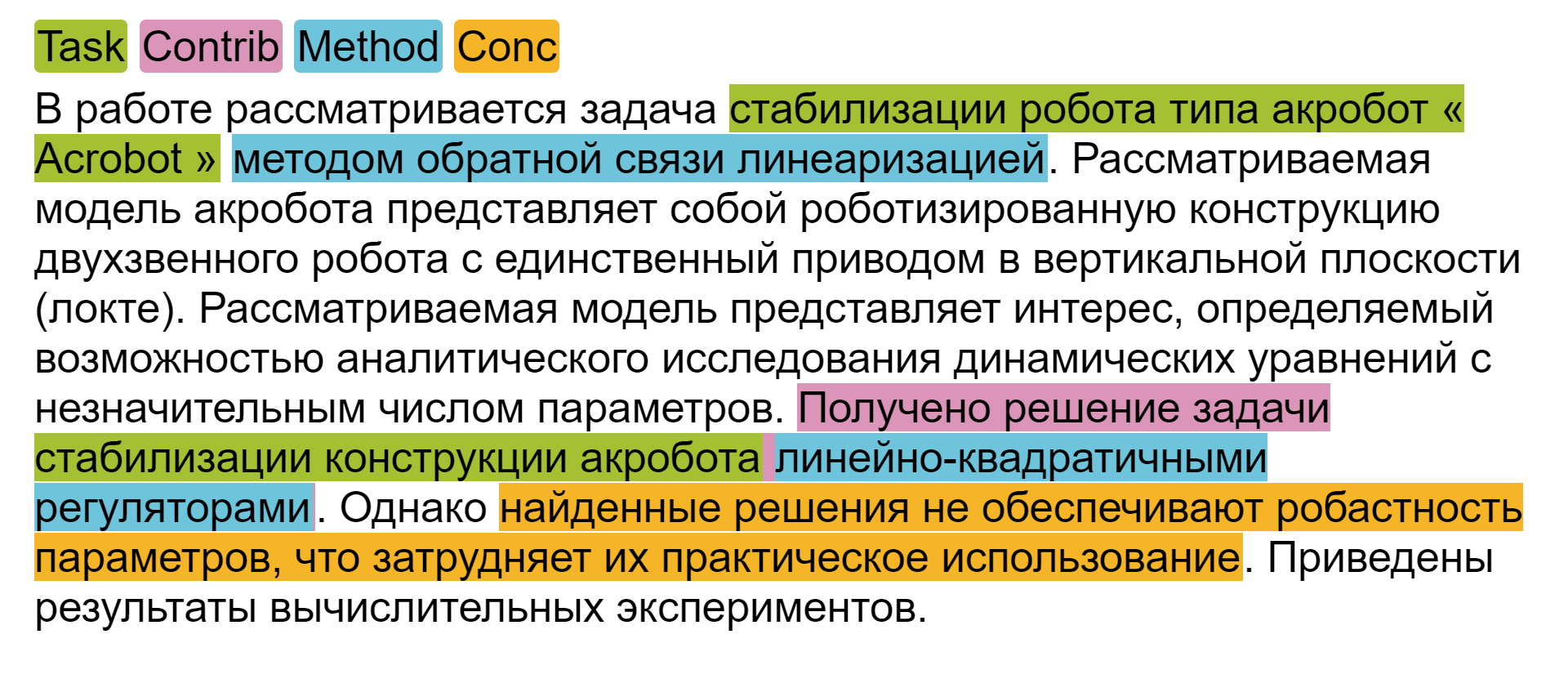}
    \caption{A manually annotated text.}
    \label{fig:manual_ru}
    \end{subfigure}
    \hfill
    \begin{subfigure}[t]{0.5\textwidth}
    \includegraphics[width=\textwidth]{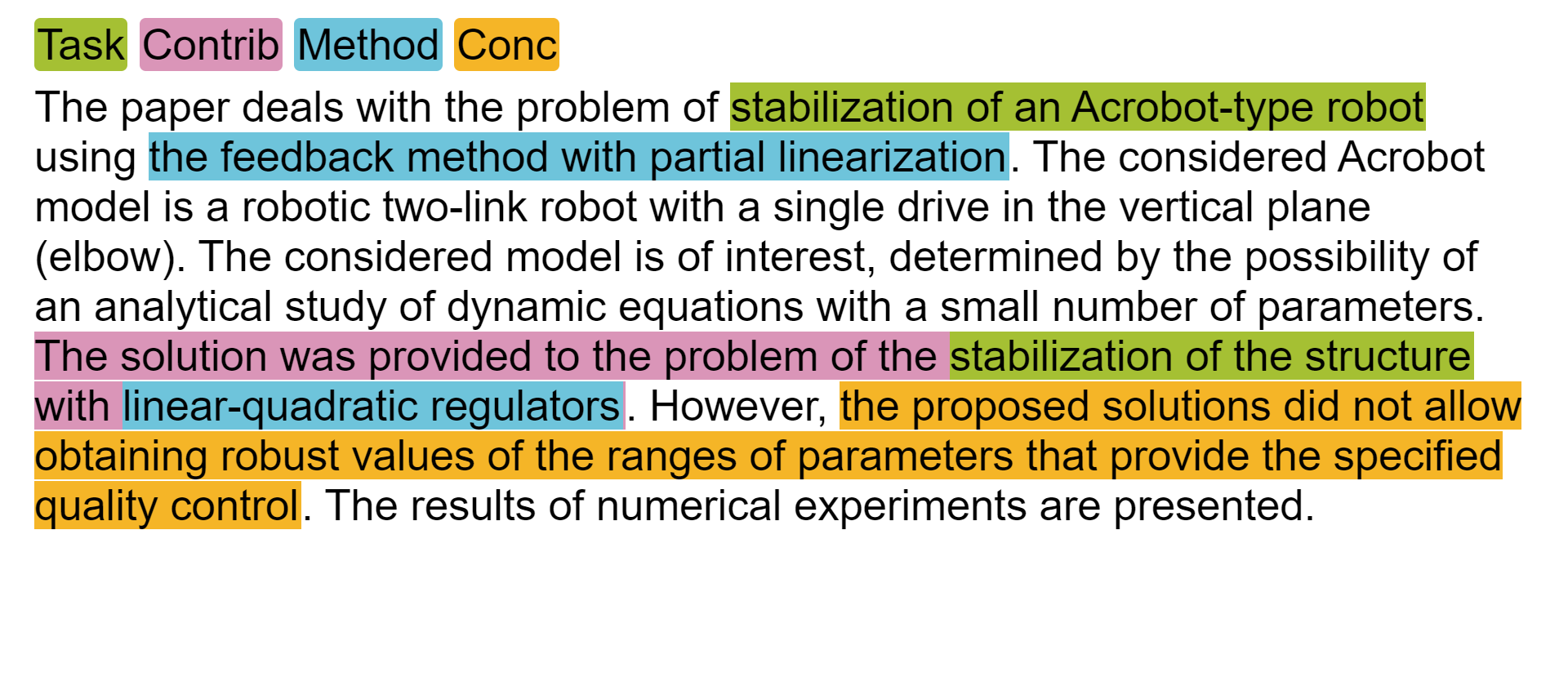}
        \vfill
    \caption{A manually annotated text, translated into English.}
    \label{fig:manual_eng}
    \end{subfigure}
    \hfill
    \begin{subfigure}[t]{0.5\textwidth}
    \centering
    \includegraphics[width=\textwidth]{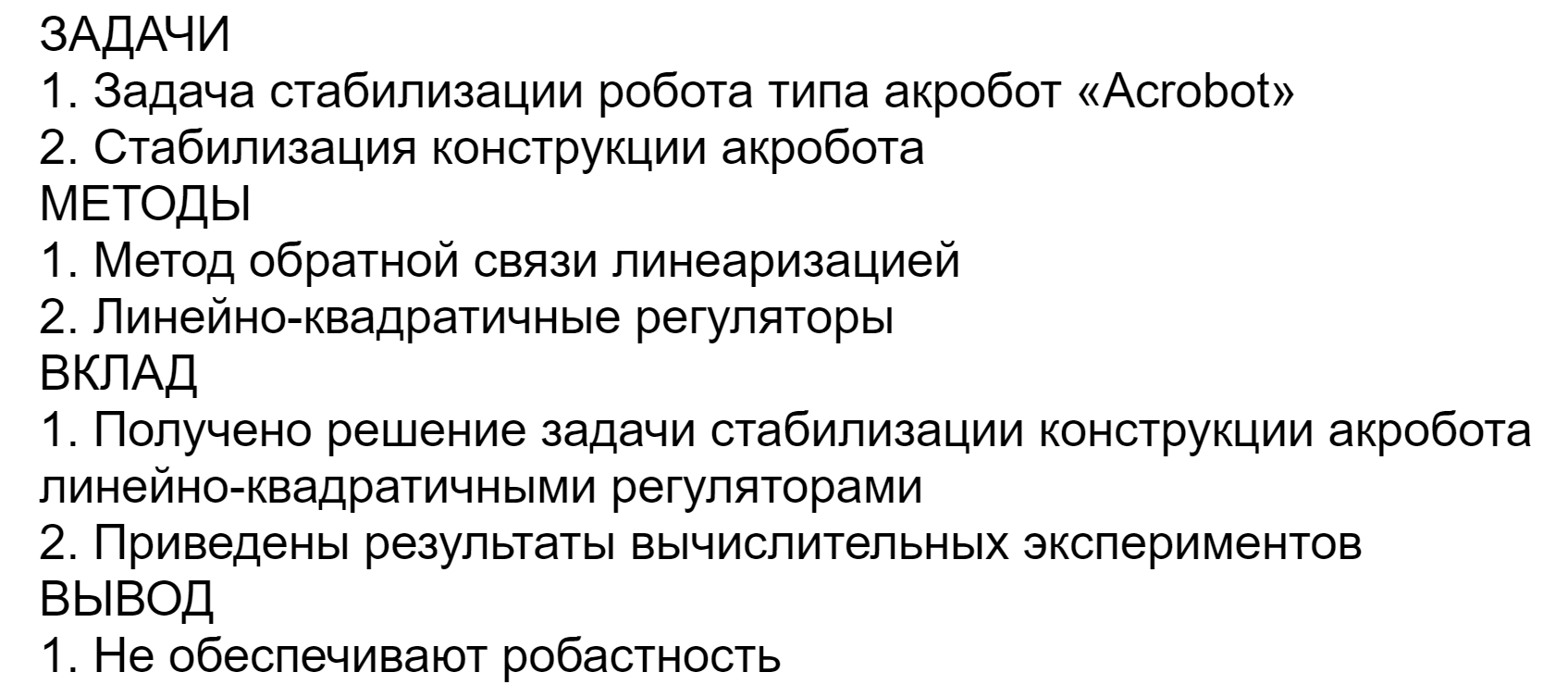}
    \caption{Aspects extracted from the text.}
    \label{fig:extracted_ru}
    \end{subfigure}
    \begin{subfigure}[t]{0.5\textwidth}
    \includegraphics[width=\textwidth]{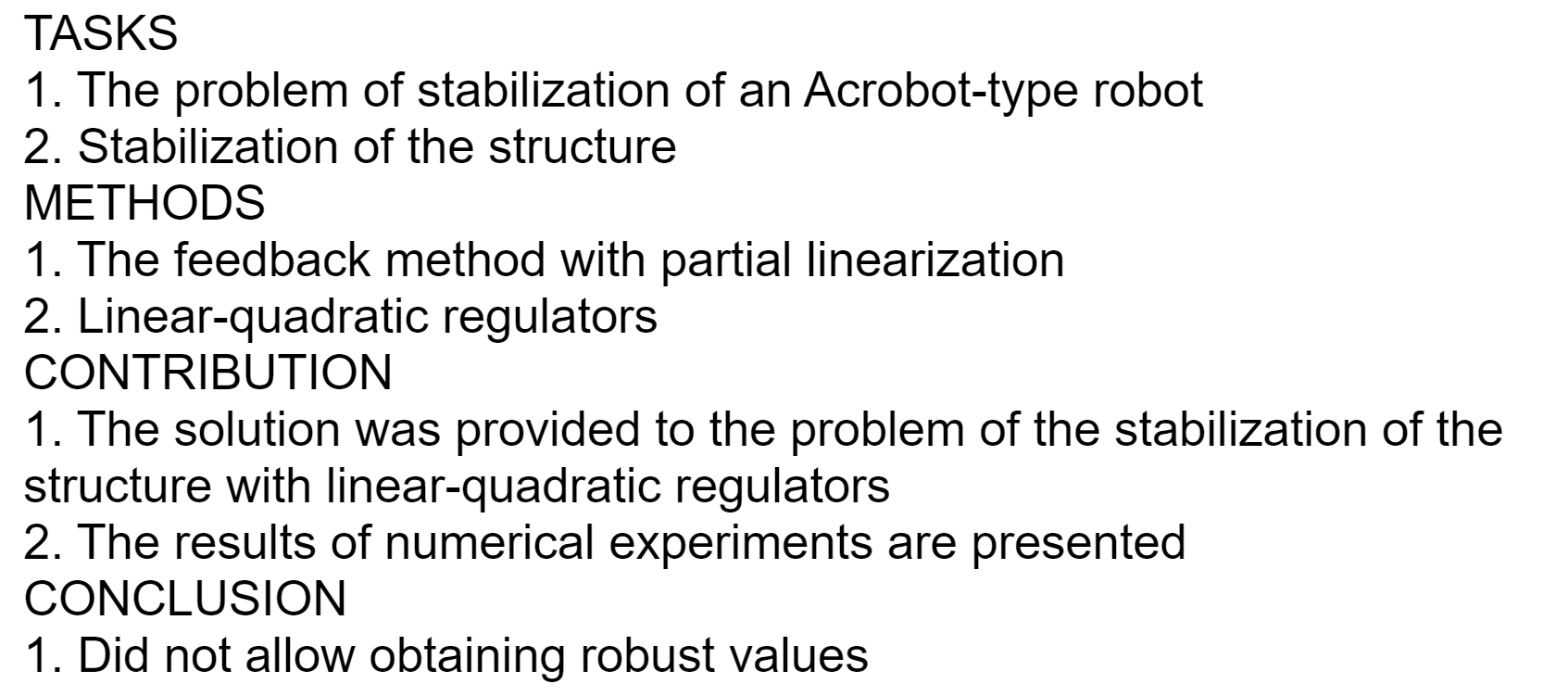}
    \caption{Aspects extracted from the text, translated into English.}
    \label{fig:extracted_eng}
    \end{subfigure}
    \caption{Example of automatic aspect extraction compared with manual annotation.}
    \label{fig:auto_vs_manual_annotation}
    
\end{figure*}

The metrics represent how our model’s predictions coincide with manual annotation. However, although the annotation was performed according to the instructions, it still has some controversial points, which means some other variants of annotation might also be possible. Having analyzed examples of annotation produced by the model, we have noticed that some of them are still viable even if they differ from those in the dataset. For example, as shown in Fig \ref{fig:extracted_ru}, the model sometimes includes the word 'the problem of' into the Task aspect. We tend to exclude such phrases in the manual annotation, but the result produced by the model can hardly be considered wrong.

Another example is the extracted conclusion shown in Fig \ref{fig:extracted_ru}. It is shorter than expected and does not fully disclose the conclusion of the original text, but still represents its main point. This can probably be considered a minor mistake.

We believe that in our case, general plausibility of results is more important than total compliance with the annotation, but at the same time, it is more difficult to measure due to its being rather subjective. In the future, we plan to explore ways to estimate it more formally.

\section{Discussion}
\label{sec:discussion}

In this paper, we propose a tool for automatic aspect extraction from scientific texts in Russian and a cross-domain dataset for this task. 
The findings of this study have to be seen in light of the following limitations:
\begin{enumerate}
    \item \textbf{Small size of the dataset.} The created dataset contains only 200 texts due to the rather small number of annotators and the labor intensiveness of the annotation process. However, even such an amount of data was sufficient to obtain primary results. Moreover, in the future, we plan to extend the dataset with more annotated texts, probably by using semi-supervised methods.
    \item \textbf{Aspect boundary detection.} Exact match ratio is lower than the metrics for individual tokens, which means that we have to continue working on improving aspect boundary detection, e.g. by developing new heuristics.
\end{enumerate}

Nevertheless, we hope that the conducted research contributed to the existing resources and methods of information extraction from scientific texts, as the proposed algorithm and dataset are the first to be cross-domain and Russian-language-oriented at the same time.

The developed tool can potentially be used to select, summarize, and systematize scientific publications, making academic literature reviewing more convenient and efficient, while the dataset can be used for further experiments in the task of aspect extraction. Both are publicly available and, we hope, will be useful for other researchers.

\section{Conclusion}
\label{conclusion}

This work presents a cross-domain dataset of Russian-language scientific texts with manual aspect annotation as well as a tool for automatic aspect extraction from Russian scientific texts of any domain.

The created dataset contains 200 texts annotated with 4 aspects: Task, Contribution, Method, and Conclusion, which were chosen according to certain features of the texts in the dataset.

To implement the algorithm of automatic aspect extraction, we fine-tuned a pretrained masked language model on our data. The best results (Macro F1 = 0.57) were obtained with multilingual BERT combined with rule-based heuristics for post-processing. The model’s robustness to different scientific domains was proven by cross-domain experiments.


\bibliographystyle{splncs04}
\selectlanguage{english}
\bibliography{mybibliography}
\end{document}